\documentclass[10pt,twocolumn,letterpaper]{article}
\usepackage[pagenumbers]{cvpr}
\usepackage{array,multirow}
\newcommand{\methodname}{MIMORPHEUS}
\newcommand{\bipara}[2]{#2}

\makeatletter
\renewcommand{\LenToUnit}[1]{\strip@pt\dimexpr#1\relax}
\makeatother

\definecolor{cvprblue}{rgb}{0.21,0.49,0.74}
\usepackage[pagebackref,breaklinks,colorlinks,allcolors=cvprblue]{hyperref}
\def\confName{CVPR}
\def\confYear{2027}
\title{\methodname{}: Rethinking Cross-Modal Supervision\\for Human Pose Estimation from Radar Signals}
\author{%
Shizhao Zhang$^{1,2}$, Yanyan Zhang$^{1,2}$, Deshui Yu$^{1,2}$, Jun Pan$^{1,3,*}$, Guangyou Fang$^{1,3,*}$\\
{$^{1}$GBA Branch of Aerospace Information Research Institute, Chinese Academy of Sciences}\\
{$^{2}$School of Electronic, Electrical, and Communication Engineering,}\\
{University of Chinese Academy of Sciences}\\
{$^{3}$Guangdong Provincial Key Laboratory of Terahertz Quantum Electromagnetics}\\[2pt]
{\small\ttfamily \{zhangshizhao24, zhangyanyan23, yudeshui22\}@mails.ucas.ac.cn, \{panjun, fanggy\}@aircas.ac.cn}
}
\newcommand{\authorcontactfootnote}{%
\begingroup
\renewcommand{\thefootnote}{\fnsymbol{footnote}}%
\footnotetext[1]{Corresponding authors: Guangyou Fang and Jun Pan.}%
\endgroup
}

\begin{document}
\maketitle
\raggedbottom
\begin{abstract}
\bipara{abs01}{Deep learning has shown promise for estimating human pose with multiple-input multiple-output (MIMO) radar through walls. Building paired radar and pose datasets demands considerable time and manual effort, while large visual pose datasets provide no corresponding radar echoes. Physical simulation can supply synthetic signals, but requires expertise in scattering models and can leave a gap between simulated and real data. To address this challenge, we introduce \methodname{}, a framework that generates radar echoes from 3D human poses through conditional diffusion. To prepare echo and pose representations, MIMO-VAE compresses the echoes according to their channel and fast time sample structure, while SkeleCoder learns pose features through contrastive alignment with echo features. These pose features then guide diffusion in the MIMO-VAE latent space, and the VAE decoder reconstructs echoes from the generated latents. Once trained, the generator augments radar data by producing different echoes for a recorded pose and supplies the radar modality for pose annotations in existing visual datasets. In data augmentation experiments, our method reduces MPJPE by 6.8\%--11.2\% across four radar pose estimators. Transfer experiments with three public pose datasets also show that visual pose annotations can supply useful radar training data. Code will be made publicly available on GitHub.}
\end{abstract}

\par
\authorcontactfootnote
\section{Introduction}
\label{sec:introduction}
\begin{figure}[t]
\centering
\includegraphics[width=\linewidth]{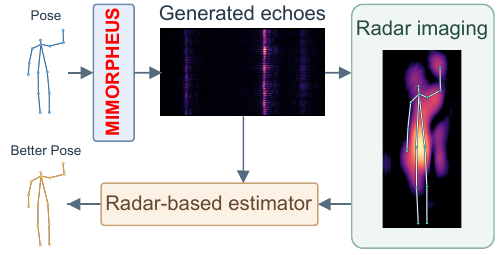}
\caption{\bipara{fig01}{\methodname{} generates complex echoes from poses and random noise. The generated pairs support pose estimation directly from echoes or after radar imaging. The framework provides a new strategy for radar data augmentation and a potential route for cross-modal data transfer.}}
\label{fig:overview}
\end{figure}

\bipara{in01}{MIMO radar can sense human bodies through many building materials without illumination. Its multiple transmit and receive antennas capture radio frequency (RF) echoes with complementary spatial information, enabling research on human pose estimation~\cite{Zhao_2018_CVPR,10.1145/3230543.3230579}. One approach converts echoes into radar images before using a neural network to estimate pose. For example, the backprojection (BP) imaging method compensates propagation delays and coherently accumulates channel echoes at candidate locations for subsequent imaging-based pose reconstruction~\cite{rs13020241}. Once range profiles are available, direct accumulation over $M$ channels and $V$ voxels costs $O(MV)$ when interpolation takes constant time. Finer grids increase $V$, making dense 3D reconstruction expensive. Although these images suit CNN backbones, converting complex reconstructions to intensity images may discard echo phase information. These limitations motivate methods that bypass imaging and estimate pose directly from echoes~\cite{10260326}.}

\bipara{in02}{Both imaging-based and direct echo pose estimators rely on large training datasets. Building such data requires synchronizing radar with cameras, registering coordinate systems, and associating pose labels with radar echoes~\cite{Choi_2025_CVPR,Fan_2026_CVPR}. Covering diverse motions, subjects, and positions takes substantial effort, which must be repeated when collecting data under new radar settings or environments. Optical datasets offer a much larger supply of pose and motion annotations~\cite{6682899,mono-3dhp2017,Joo_2015_ICCV,Li_2021_ICCV,Wang_2024_CVPR}. However, these resources lack corresponding radar echoes and therefore cannot be used directly to train radar pose estimators.}

\bipara{in03}{Simulation provides another way to generate radar training data. Motion can drive scattering and signal models to synthesize echoes. SynMotion already reuses skeletons from visual datasets to produce labeled raw millimeter-wave (mmWave) signals through explicit physical models~\cite{10.1145/3560905.3568542}. Such approaches require expertise in body geometry, antenna configuration, and propagation. Detailed simulation across many 3D body configurations can also become computationally demanding at the scale needed for deep learning. Differences between simulated and measured echoes can limit model generalization~\cite{10.1145/3625687.3625798}.}

\bipara{in04}{Learned synthesis addresses some of these limitations. RF Genesis combines physical simulation with generative refinement to synthesize radar analog-to-digital converter (ADC) data, but still requires explicit scene and propagation models~\cite{10.1145/3625687.3625798}. mmWaveFlow generates or enhances human radar point clouds, which cannot directly supply the complex echoes required by echo models~\cite{Su_2026_CVPR}. These studies expand radar training data through synthesis. However, augmenting complex echoes remains challenging because changes in body position or orientation affect propagation paths and phase across channels. Simple spatial transformations used for optical images therefore do not generally yield valid new echoes. We seek a method that generates complex echoes from pose annotations for data augmentation, without constructing a physical scene for each pose.}

\bipara{in05}{To address these limitations, we train a complex echo generator on paired radar echoes and poses for data augmentation. It is important for the development of deep learning in radar sensing. In addition, we find that existing pose annotations can be used to generate missing radar echoes, providing a strategy for cross-modal data transfer. As shown in Figure~\ref{fig:overview}, once trained, \methodname{} can sample additional echoes for recorded poses or accept mapped poses from external visual datasets. The former augments measured radar data, while the latter supplies paired radar echoes for public pose datasets.}

\bipara{in06}{In this paper, we introduce \methodname{}, a new framework for generating multichannel complex MIMO echoes from 3D skeletons through conditional diffusion. We address the structural difference between joint coordinates and echo signals by learning representations for both modalities. MIMO-VAE compresses multichannel echoes while accounting for virtual channel structure and the ordering of real and imaginary samples. SkeleCoder learns joint relations and is pretrained by aligning global skeleton features with global radar latent features, providing pose features for conditional generation. The diffusion model combines these features with body geometry to generate latents, which the MIMO-VAE decoder converts to complex echoes. Once trained, \methodname{} generates echoes paired with the input poses, without additional acquisition or scene specification. We evaluate whether these generated pairs provide useful supervision through radar data augmentation and transfer from public visual pose datasets. Augmentation reduces MPJPE by 6.8\%--11.2\% across four radar pose estimators on measured test echoes. Transfer experiments with three public pose datasets further show that estimators trained on the generated pairs can recover poses from measured radar echoes.}

\bipara{in07}{The contributions of this paper are summarized as follows. (1) We introduce \methodname{}, a new strategy for radar data augmentation and cross-modal data transfer through pose-conditioned echo generation. It produces additional echoes for recorded poses and supplies the missing radar modality for public visual pose datasets, extending radar supervision beyond newly collected pairs. (2) We design MIMO-VAE, a MIMO radar signal compression model guided by the physical structure of echoes. It accounts for correlations across virtual channels and neighboring fast-time samples, and retains paired real and imaginary components that jointly encode amplitude and phase, providing compact radar representations for diffusion. (3) We develop SkeleCoder and a contrastive strategy that aligns global skeleton features with radar latent features. This connects the two modalities without imposing a correspondence between joints and radar sampling positions, supporting stable training of the subsequent conditional diffusion model.}

\section{Related work}
\label{sec:related}
\subsection{Learning human pose from RF signals}
\bipara{rw01}{Learning methods recover human structure from radar images, Doppler signatures, or complex echoes. RF-Pose uses visual supervision to estimate 2D poses from radio reflections~\cite{Zhao_2018_CVPR}. RF-Pose3D extends recovery to 3D skeletons and decomposes high-dimensional convolutions to process spatial and temporal information~\cite{10.1145/3230543.3230579}. For through-wall pose reconstruction, Song et al. combine BP imaging with a 3D CNN~\cite{rs13020241}. RadarFormer bypasses imaging and processes echoes with Transformers, improving accuracy and reducing computation in its evaluation~\cite{10260326}. MVDoppler-Pose combines multiple views and Doppler cues to improve walking pose estimation at long distances and under self-occlusion~\cite{Choi_2025_CVPR}. These studies improve human reconstruction through architecture design and feature fusion, but still depend on substantial data collected for their sensing setups. We therefore focus on training data rather than further architectural changes to the pose estimator.}

\subsection{Simulation and synthesis}
\bipara{rw02}{Simulation generates radar training data without recording each motion again. Vid2Doppler synthesizes Doppler signatures from video, while SimHumalator provides motion-driven radar simulation~\cite{10.1145/3411764.3445138,9664284}. RadSimReal addresses the gap between simulated and measured radar data for object detection~\cite{Bialer_2024_CVPR}. MoCap2Radar instead predicts micro-Doppler signatures from motion capture~\cite{Chen2025MoCap2RadarAS}. Applying these methods to different tasks and scenes can require domain expertise in physical models, radar settings, or output representations. Differences between simulated and measured data may also limit generalization.}

\subsection{Generative models for data}
\bipara{rw03}{Diffusion models learn data distributions through denoising and have been applied to conditional image synthesis, with recent studies improving generation quality and diversity~\cite{Ho2020DenoisingDP,Rombach_2022_CVPR,chen2026taming,hu2026blameerrantwordscanvas}. Such models can also supply training data for perception, as demonstrated by AeroGen for remote sensing detection~\cite{Tang_2025_CVPR}. In radar, the LiDAR-to-4DRadar diffusion bridge aligns latent representations to translate light detection and ranging (LiDAR) data into radar tensors~\cite{Shen_2026_CVPR}. We focus on MIMO radar and use diffusion to generate echoes that follow the measured data distribution. This adds a data augmentation method for MIMO radar and offers a new approach to cross-modal data transfer. Recent work on radar foundation models further highlights the importance of data scale~\cite{Huang_2025_ICCV}. We believe that addressing data limitations can substantially advance this field.}

\section{Proposed method}
\label{sec:method}
\bipara{me00}{This section describes the design of \methodname{}. We first define the data and task, then introduce MIMO-VAE, the contrastive pretraining of SkeleCoder, and finally the operation of the conditional diffusion model.}

\subsection{Acquisition and problem formulation}
\bipara{me01}{Our platform combines a frequency-modulated continuous-wave (FMCW) radar, a color and depth (RGB-D) camera, and a computer. The radar array provides $M=100$ virtual channels. A visual model estimates 14 landmarks from the camera data, comprising the nose, neck, and paired shoulders, elbows, wrists, hips, knees, and ankles. We align the radar and camera coordinate systems using a reflector target~\cite{Cheng20233DRA}. Computer timestamps align the two streams. Each pair contains an echo represented as $X\in\mathbb R^{M\times N\times2}$ and a skeleton $P\in\mathbb R^{J\times3}$, where $N=1023$ is the number of fast-time samples, $J=14$ is the number of joints, and the last dimension of $X$ stores the real and imaginary (RI) components.}

\bipara{me02}{\methodname{} learns the conditional distribution $p_\theta(X\mid P)$. Once trained, it generates an echo from a pose and a random seed. Different seeds produce additional echoes for radar augmentation, while skeletons from vision datasets supply labels. Training proceeds through echo compression, skeleton--radar alignment, and conditional generation.}

\subsection{MIMO-VAE: Complex echo representation}
\label{sec:mimovae}
\begin{figure}[t]
\centering
\includegraphics[width=\linewidth]{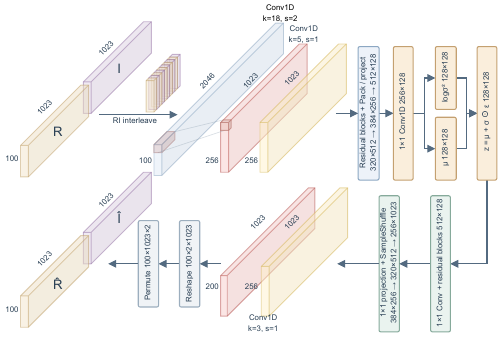}
\caption{\bipara{fig02}{MIMO-VAE. Guided by the physical properties of echoes, the first convolution processes complete complex fast-time samples and mixes virtual channels. Packing and sample shuffle organize compression and expansion along the fast-time axis. Learned projections change feature width.}}
\label{fig:mimovae}
\end{figure}

\bipara{me03}{Direct diffusion over the input echoes requires denoising a large number of samples. MIMO-VAE reduces this cost by learning a compact representation from which the echoes can be reconstructed. We use BP imaging to explain the physical meaning of complex echoes and motivate the representation design. Figure~\ref{fig:mimovae} shows the network design.}

\bipara{me03_bp}{BP divides a 3D region into voxels and coherently accumulates the channel responses at each candidate position $\mathbf x$~\cite{rs13020241}. A discrete formulation is}
\begin{equation}
\begin{aligned}
\tau_m(\mathbf x)&=\frac{\|\mathbf x-\mathbf p_m^{\mathrm{Tx}}\|+\|\mathbf x-\mathbf p_m^{\mathrm{Rx}}\|}{c},\\
b_m(\mathbf x)&=\sum_{n=0}^{N-1}Y_m[n]K_m(t_n,\tau_m(\mathbf x)),\\
\mathcal I_{\mathrm{BP}}(\mathbf x)&=\left|\sum_{m=1}^{M}b_m(\mathbf x)\right|^2.
\end{aligned}
\label{eq:bp_intensity}
\end{equation}
\bipara{me03_physics}{Here, $Y_m[n]=X_{m,n,0}+\mathrm iX_{m,n,1}$ is the complex echo at fast-time sample $t_n=n\Delta t$. The delay $\tau_m(\mathbf x)$ is determined by the transmit and receive antenna positions $\mathbf p_m^{\mathrm{Tx}}$ and $\mathbf p_m^{\mathrm{Rx}}$ for virtual channel $m$, and propagation speed $c$. The focusing kernel $K_m$ includes range focusing and phase compensation for the acquisition waveform. At a scatterer, the compensated responses tend to add constructively; at other positions, they are less coherent. Neighboring fast-time samples belong to the same ordered waveform and share local signal structure. In contrast, virtual channel indices identify antenna pairs and do not by themselves define a uniform 1D neighborhood. We therefore apply local convolutions along fast time and jointly mix the virtual channels.}

\bipara{me03_ri}{The image intensity also depends jointly and nonlinearly on the real and imaginary contributions. MIMO-VAE therefore retains both components as paired features, interleaving them as $[I_0,Q_0,I_1,Q_1,\ldots]$, so convolutions and nonlinear activations can learn local functions of the paired values. The first convolution uses an even kernel, with stride and padding aligned to complete RI pairs. It jointly mixes virtual channels, whose responses describe the same body from different antenna pairs. Subsequent convolutions refine these features before compression.}

\bipara{me04}{The encoder gradually increases feature width as it reduces fast-time resolution. Residual blocks with group normalization (GroupNorm), sigmoid-weighted linear unit (SiLU) activations, and convolutions refine features at each scale~\cite{He_2016_CVPR,Wu_2018_ECCV,ELFWING20183}. Between scales, packing rearranges adjacent fast-time positions into feature channels before a learned projection reduces their width. This packing is a 1D inverse of sub-pixel rearrangement~\cite{Shi_2016_CVPR}. The rearrangement itself preserves values, with right padding for an odd length. The decoder uses learned projections and sample shuffle to restore the fast-time layout without interpolation. Only the packing and shuffle operations are inverse rearrangements, while compression and reconstruction are learned. The output is reorganized into RI pairs for each virtual channel.}

\bipara{me05}{The posterior head predicts a mean and log variance that define a diagonal Gaussian $q_\phi(u\mid X)$~\cite{Kingma2013AutoEncodingVB}. During variational autoencoder (VAE) training, we sample $u=\mu_\phi(X)+\sigma_\phi(X)\odot\epsilon$, where $\epsilon\sim\mathcal N(0,I)$, and reconstruct $\widehat X=D_\phi(u)$. We optimize}
\begin{equation}
\begin{aligned}
\mathcal L_{\mathrm{VAE}}={}&\mathcal L_{\mathrm{RI}}+0.2\mathcal L_{\mathrm{amp}}
+0.1\mathcal L_{\mathrm{NMSE}}\\
&+\beta D_{\mathrm{KL}}\!\left(q_\phi(u\mid X)\,\|\,\mathcal N(0,I)\right),
\end{aligned}
\label{eq:vae}
\end{equation}
\bipara{me06}{where $\mathcal L_{\mathrm{RI}}$ is the mean absolute error of the real and imaginary components. The amplitude term compares $\log(1+|\widehat Y|)$ and $\log(1+|Y|)$, where $Y=X_{:,:,0}+\mathrm iX_{:,:,1}$ and $\widehat Y$ is the complex reconstruction corresponding to $\widehat X$. The normalized mean squared error (NMSE) averages $\|\widehat X-X\|_2^2/\max(\|X\|_2^2,\varepsilon)$ over samples. These terms constrain component errors, amplitude structure, and relative reconstruction error. The Kullback--Leibler (KL) term is averaged over latent elements, with $\beta$ increased gradually during warmup. Alignment and diffusion use the posterior mean to avoid random variation in their radar targets. Diffusion scales it as $z_0=s\mu_\phi(X)$, where $s$ is the inverse root mean square (RMS) of training posterior means.}

\subsection{SkeleCoder: Representation alignment}
\label{sec:alignment}
\bipara{me07_init}{We introduce SkeleCoder to encode joint coordinates for diffusion conditioning. Our ablation (Section~\ref{sec:ablations}) shows that joint training from random initialization yields less useful echoes for augmentation. We therefore prealign global skeleton and radar features in a shared embedding space through contrastive learning inspired by Contrastive Language--Image Pretraining (CLIP)~\cite{pmlr-v139-radford21a}. This initializes SkeleCoder for stable joint optimization with diffusion, during which it continues to update.}

\begin{figure}[t]
\centering
\includegraphics[width=\linewidth]{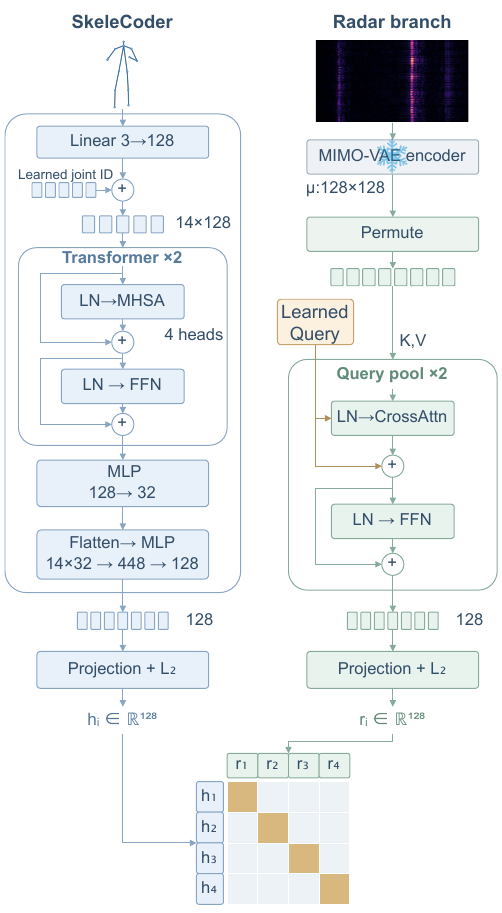}
\caption{\bipara{fig03}{Skeleton--radar alignment with SkeleCoder. SkeleCoder retains joint order during skeleton aggregation, while a learned query reads radar fast-time tokens derived from the frozen MIMO-VAE posterior mean. The loss aligns global features from the two modalities. The full joint tokens remain available for diffusion conditioning.}}
\label{fig:alignment}
\end{figure}

\begin{figure*}[t]
\centering
\includegraphics[width=\linewidth]{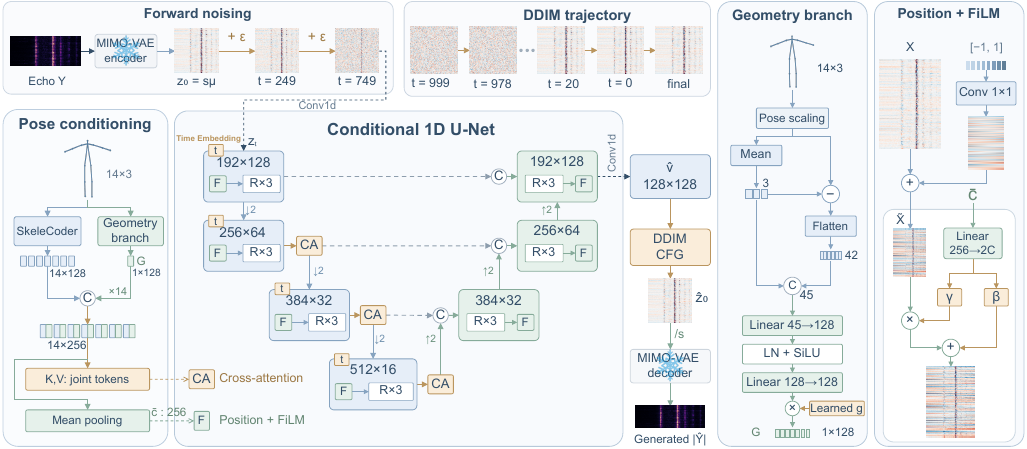}
\caption{\bipara{fig04}{Conditional 1D latent generation. Training latents come from the frozen MIMO-VAE mean, followed by scaling and noise addition. SkeleCoder joint and geometry features are concatenated before conditioning. FiLM modulates features at every scale, while radar queries read the condition tokens through cross-attention at the coarser encoder scales. Diffusion generates a latent that the frozen MIMO-VAE decoder converts to complex echoes.}}
\label{fig:diffusion}
\end{figure*}

\bipara{me07_context}{Joints and radar samples have different physical meanings and no one-to-one correspondence. Joint positions are spatially related, while radar samples have dependencies along fast time. We therefore use Transformer attention to aggregate context within each modality before aligning their global features~\cite{Vaswani2017AttentionIA}. SkeleCoder embeds joint coordinates and adds learned joint identities. Spatial Transformer layers model relations among joints while retaining a separate token for each joint. A shared multilayer perceptron (MLP) compresses each token, then an MLP aggregates the flattened tokens in their fixed joint order. This aggregates the ordered joint features without an explicit average over joints, while the full joint tokens remain available for diffusion.}

\bipara{me08}{After training, MIMO-VAE remains frozen. The radar branch treats the frozen posterior mean as tokens ordered along the compressed fast-time axis. A learned query gathers their information through successive cross-attention blocks with residual connections and feedforward layers. Each block refines the query using the same radar context. RoPE provides position cues along this axis~\cite{10.1016/j.neucom.2023.127063}. Separate projection MLPs map radar and skeleton global features to unit vectors $r_i$ and $h_i$ in a shared embedding space. The radar query and projections are learned together with SkeleCoder.}

\bipara{me09}{For a global batch of $B$ pairs and learned logit scale $\kappa$, we optimize the following symmetric contrastive objective:}
\begin{equation}
\ell(r,h)=-\frac1B\sum_{i=1}^{B}
\log\frac{\exp(\kappa r_i^\top h_i)}
{\sum_{j=1}^{B}w_{ij}\exp(\kappa r_i^\top h_j)},
\label{eq:clip_direction}
\end{equation}
\begin{equation}
\mathcal L_{\mathrm{align}}=\tfrac12\bigl[\ell(r,h)+\ell(h,r)\bigr].
\label{eq:alignment}
\end{equation}
\bipara{me10}{Similar poses in a continuous recording can otherwise become false negatives. We set $w_{ii}=1$ and use}
\begin{equation}
\begin{aligned}
w_{ij}&=1-(1-w_{\min})\exp(-d_{ij}^2/2)\,a_{ij},\\
a_{ij}&=\tfrac12+\tfrac12\mathbb{I}[g_i=g_j]
\exp(-|\tau_i-\tau_j|/\sigma_t),
\end{aligned}
\label{eq:negative}
\end{equation}
\bipara{me11}{for other pairs. The distance $d_{ij}^2$ sums normalized squared distances between centered joint layouts and between body centers. The recording identifier $g_i$ and timestamp $\tau_i$ describe acquisition context. The indicator $\mathbb{I}[g_i=g_j]$ equals 1 for samples from the same recording and 0 otherwise. We use $w_{\min}=0.25$ and $\sigma_t=1$~s. Pose similarity reduces the negative penalty, and temporal proximity within a recording strengthens that reduction. Timestamps enter only the loss. This objective aligns global pose and echo features without forcing joint tokens to reproduce the radar fast-time grid. Figure~\ref{fig:alignment} shows the alignment architecture.}

\subsection{Pose-conditioned latent diffusion}
\label{sec:diffusion}
\bipara{me12_intro}{MIMO-VAE provides compact echo latents, while SkeleCoder supplies pose features for generation. The latent retains an ordered axis inherited from fast-time sampling. We use a 1D U-Net to model dependencies along this axis at multiple resolutions, combining noisy echo latents with pose features through cross-attention and feature-wise linear modulation (FiLM). Figure~\ref{fig:diffusion} summarizes the generation architecture.}

\bipara{me12}{We use SkeleCoder's joint tokens $H_j$ as one source of conditioning. To supply explicit cues about body layout and position, we add a geometry branch that encodes centered joints and the body center. For a fixed body scale $d$, let $c=\frac1J\sum_jP_j/d$ and $R_j=P_j/d-c$. An MLP maps $[\operatorname{vec}(R);c]$ to a global feature $G$, with a learned gain initialized to 0.1. We broadcast $G$ across joints and concatenate it with each joint token as $\mathcal C_j=[H_j;G]$, where $j$ indexes the $J$ joints.}

\bipara{me13}{Concatenation retains separate feature groups for joint relations and global geometry before learned condition projections. We model echo latents with conditional latent diffusion~\cite{Rombach_2022_CVPR}. A cosine schedule controls noise addition~\cite{pmlr-v139-nichol21a}. With $a_k=\sqrt{\bar\alpha_k}$, $b_k=\sqrt{1-\bar\alpha_k}$, and a uniformly sampled diffusion step $k$, we form}
\begin{equation}
z_k=a_kz_0+b_k\epsilon,\qquad v_k=a_k\epsilon-b_kz_0.
\label{eq:diffusion}
\end{equation}
\bipara{me14}{A 1D U-Net predicts the velocity target $v_k$, a weighted combination of noise and the clean latent~\cite{salimans2022progressive}. Convolutional residual blocks operate at several fast-time resolutions. Downsampling expands their receptive field, while skip connections carry finer features to the decoder. A learned diffusion-step embedding identifies the noise level. At coarser encoder scales, cross-attention lets radar features query the joint conditions $\mathcal C$. RoPE encodes radar positions, and SkeleCoder supplies joint identities. To also condition features at every resolution on the overall body configuration, FiLM maps the mean condition $\overline{\mathcal C}$ to channelwise parameters $(\gamma,\delta)$ and modulates features as $F\odot(1+0.1\tanh\gamma)+0.1\delta$~\cite{Perez_Strub_de_Vries_Dumoulin_Courville_2018}. The resulting channelwise scaling and shifts propagate global pose information to both coarse and fine features. Learned projections of normalized fast-time coordinates provide additional position cues.}

\bipara{me15}{MIMO-VAE and the radar alignment branch remain frozen. We jointly update the denoiser, SkeleCoder, skeleton projection, and geometry branch using}
\begin{equation}
\mathcal L=\mathbb E\|v_\theta(z_k,k,\mathcal C)-v_k\|_2^2
+0.05\mathcal L_{\mathrm{align}}+0.05\mathcal L_{\mathrm{str}}.
\label{eq:objective}
\end{equation}
\bipara{me16}{All elementwise losses are averaged. Alongside velocity prediction and alignment, $\mathcal L_{\mathrm{str}}$ constrains echo strength and energy location along fast time. It compares $D_\phi(\widehat z_0/s)$ with $D_\phi(z_0/s)$, where $\widehat z_0=a_kz_k-b_kv_\theta$. Weighted log-amplitude error emphasizes strong reference samples, cumulative normalized power profiles constrain energy location, and an error on $\log(1+\operatorname{RMS})$ for each channel constrains echo strength. Their weights are $1,0.5,0.5$. We evaluate this term on at most two samples per graphics processing unit (GPU), retaining only samples with pose conditions and diffusion steps $k\leq500$. Gradients pass through the frozen decoder to the generator.}

\bipara{me17}{Training replaces the entire pose condition with a learned null condition with probability 0.1 for classifier-free guidance~\cite{Ho2022ClassifierFreeDG}. Generation uses the denoising diffusion implicit model (DDIM)~\cite{Song2020DenoisingDI}, combining conditional and unconditional predictions as $v_{\mathrm u}+2(v_{\mathrm c}-v_{\mathrm u})$. The final latent is divided by $s$, decoded to complex echoes, and restored to the radar RMS scale. A fixed joint mapping and coordinate conversion express public 3D poses in the same skeleton convention. Generated echoes retain those poses as labels and can train either direct echo models or models that first reconstruct radar images.}

\section{Evaluating radar data generation}\label{sec:experiments}
\bipara{ex01}{We evaluate \methodname{} by augmenting a measured radar dataset and transferring poses from established vision datasets into radar training pairs. A controlled comparison examines how alignment pretraining affects the utility of generated echoes for data augmentation.}

\begin{table}[t]
\centering
\caption{\bipara{tab01}{Radar data augmentation. MPJPE (mm) is evaluated on the measured test set using the same 14 target joints. $\dagger$ denotes reimplementation from the paper where official code was unavailable.}}
\label{tab:augmentation}
\small
\setlength{\tabcolsep}{2.5pt}
\begin{tabular}{@{}llc@{}}
\toprule
Estimator & Training data & MPJPE $\downarrow$\\
\midrule
RadarFormer~\cite{10260326} & Real & $41.3$\\
 & Real + \methodname{} & $38.5$\\
RF-Pose3D$^{\dagger}$~\cite{10.1145/3230543.3230579} & Real & $43.9$\\
 & Real + \methodname{} & $39.1$\\
BP + 3D CNN$^{\dagger}$~\cite{rs13020241} & Real & $41.2$\\
 & Real + \methodname{} & $36.6$\\
RFMamba$^{\dagger}$~\cite{DBLP:conf/iclr/ZhangGLSGZ0025} & Real & $55.8$\\
 & Real + \methodname{} & $51.3$\\
\bottomrule
\end{tabular}
\end{table}

\begin{table*}[t]
\centering
\caption{\bipara{tab02}{Transfer from public pose datasets. MPJPE (mm) is reported on generated source echoes and measured radar echoes.}}
\label{tab:transfer}
\small
\setlength{\tabcolsep}{2.5pt}
\begin{tabular}{@{}lcccc@{}}
\toprule
 & \multicolumn{2}{c}{Source-generated MPJPE $\downarrow$} & \multicolumn{2}{c}{Real-target MPJPE $\downarrow$}\\
\cmidrule(lr){2-3}\cmidrule(l){4-5}
Pose source & RF-P3D-SF$^{\dagger}$ & BP + 3D CNN$^{\dagger}$ & RF-P3D-SF$^{\dagger}$ & BP + 3D CNN$^{\dagger}$\\
\midrule
CMU Panoptic~\cite{Joo_2015_ICCV} & $41.7$ & $39.1$ & $39.3$ & $38.6$\\
AIST++~\cite{Li_2021_ICCV} & $41.9$ & $38.0$ & $36.6$ & $35.8$\\
FreeMan~\cite{Wang_2024_CVPR} & $45.1$ & $43.9$ & $43.5$ & $42.3$\\
\bottomrule
\end{tabular}
\end{table*}

\subsection{Experimental setup}
\bipara{ex02}{We collected two datasets containing 180,420 and 82,270 paired echo and pose frames, denoted A and B, respectively, with balanced coverage of positions, actions, body builds, and genders.{\renewcommand{\thefootnote}{\fnsymbol{footnote}}\footnotemark[1]} Dataset A is used to train \methodname{}. Dataset B is split into training, validation, and test sets with a target ratio of 8:1:1 for the augmentation experiments with four pose estimation backbones. Training uses four RTX 3090 GPUs and AdamW. For \methodname{}, MIMO-VAE is trained for 200 epochs with an initial learning rate of $2\times10^{-4}$. Alignment pretraining runs for 50 epochs at $10^{-4}$. Diffusion is trained for 3,000 epochs, with peak learning rates of $6\times10^{-5}$ for the denoiser and $10^{-5}$ for pose conditioning. After 30 warmup epochs, cosine decay reduces these rates to 10\% of their peaks. Per-GPU batch sizes are 32, 128, and 64 for the three stages, respectively. Sampling uses 50 DDIM steps with guidance scale 2.}

{\renewcommand{\thefootnote}{\fnsymbol{footnote}}
\footnotetext[1]{\bipara{ex02_note}{All participants provided informed consent. Protective measures were implemented during data collection, and the radar transmit power was maintained at a safe, low level.}}}

\bipara{ex04}{Downstream evaluation uses mean per joint position error (MPJPE) in millimeters, defined as $\operatorname{MPJPE}=\frac{1}{LJ}\sum_{\ell=1}^{L}\sum_{j=1}^{J}\|\widehat{\mathbf p}_{\ell j}-\mathbf p_{\ell j}\|_2$, where $L$ is the number of evaluated frames, $J$ is the number of joints, and $\widehat{\mathbf p}_{\ell j}$ and $\mathbf p_{\ell j}$ are the predicted and reference joint positions in radar coordinates.}

\subsection{Radar data augmentation}\label{sec:augmentation}

\bipara{ex05}{We evaluate augmentation with RadarFormer, RF-Pose3D, the BP and 3D CNN method of Song et al., and RFMamba~\cite{10260326,10.1145/3230543.3230579,rs13020241,DBLP:conf/iclr/ZhangGLSGZ0025}. RadarFormer and RFMamba process echo sequences. RF-Pose3D receives horizontal and vertical heatmaps derived from BP reconstruction, while the 3D CNN receives the reconstructed volume. We compare augmentation against each estimator's real-only control, keeping its architecture and training budget fixed.}

\bipara{ex07}{Table~\ref{tab:augmentation} shows lower MPJPE for all four estimators after augmentation, with reductions of 2.8--4.8 mm, or 6.8\%--11.2\%, relative to training on measured data alone. The gains span different input representations and architectures, supporting the use of multiple echo samples for a fixed pose. Such samples expose an estimator to more radar input variations with the same pose label and can reduce its dependence on a particular measured echo. The imaging methods show larger relative reductions than the direct echo methods. This indicates that augmentation at the signal level can benefit pose estimation after BP reconstruction. Sampling multiple complex echoes can yield different focused intensity patterns for the same pose, providing a broader set of training inputs than a single reconstruction. These variations offer richer supervision to estimators that rely on intensity heatmaps or volumes, which discard the phase of the complex reconstruction.}

\begin{figure*}[t]
\centering
\begingroup
\footnotesize
\newcommand{\qualitativepanel}[1]{\raisebox{-0.5\height}{\includegraphics[width=1.85in]{#1}}}
\newcommand{\qualitativeposepanel}[1]{\raisebox{-0.5\height}{\includegraphics[height=1.05in,trim=50bp 99bp 16bp 60bp,clip]{#1}}}
\begin{tabular}{@{}l@{\hspace{5pt}}c@{\hspace{5pt}}c@{\hspace{5pt}}c@{}}
 & (a) Reference echo & (b) Generated echo & (c) BP volume and pose\\[4pt]
RadarFormer &
\qualitativepanel{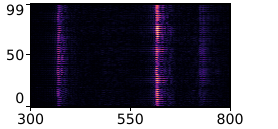} &
\qualitativepanel{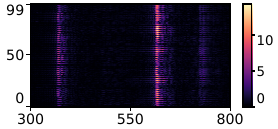} &
\qualitativeposepanel{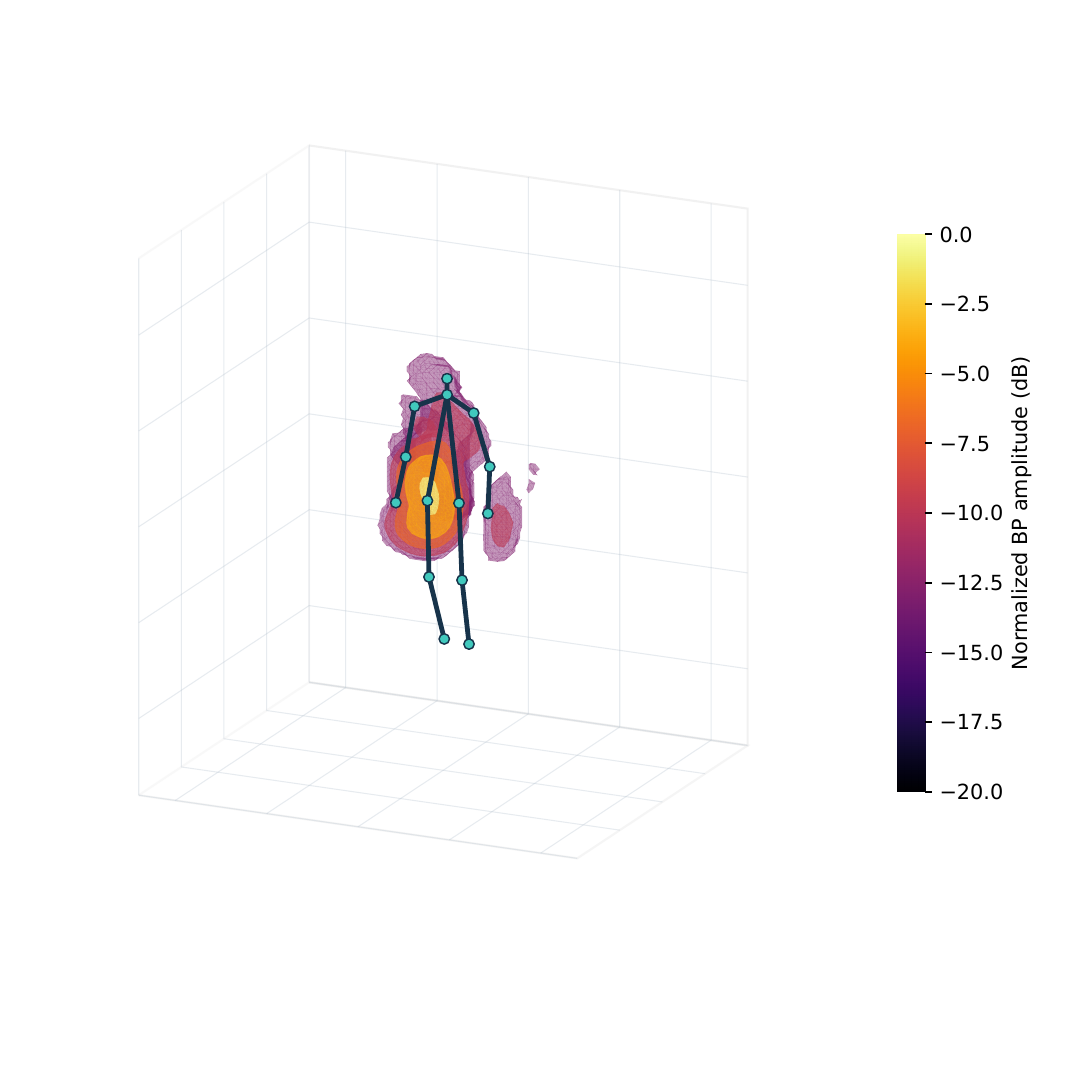}\\[4pt]
RF-Pose3D$^{\dagger}$ &
\qualitativepanel{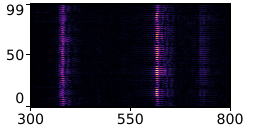} &
\qualitativepanel{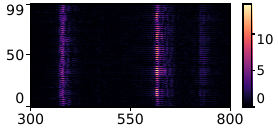} &
\qualitativeposepanel{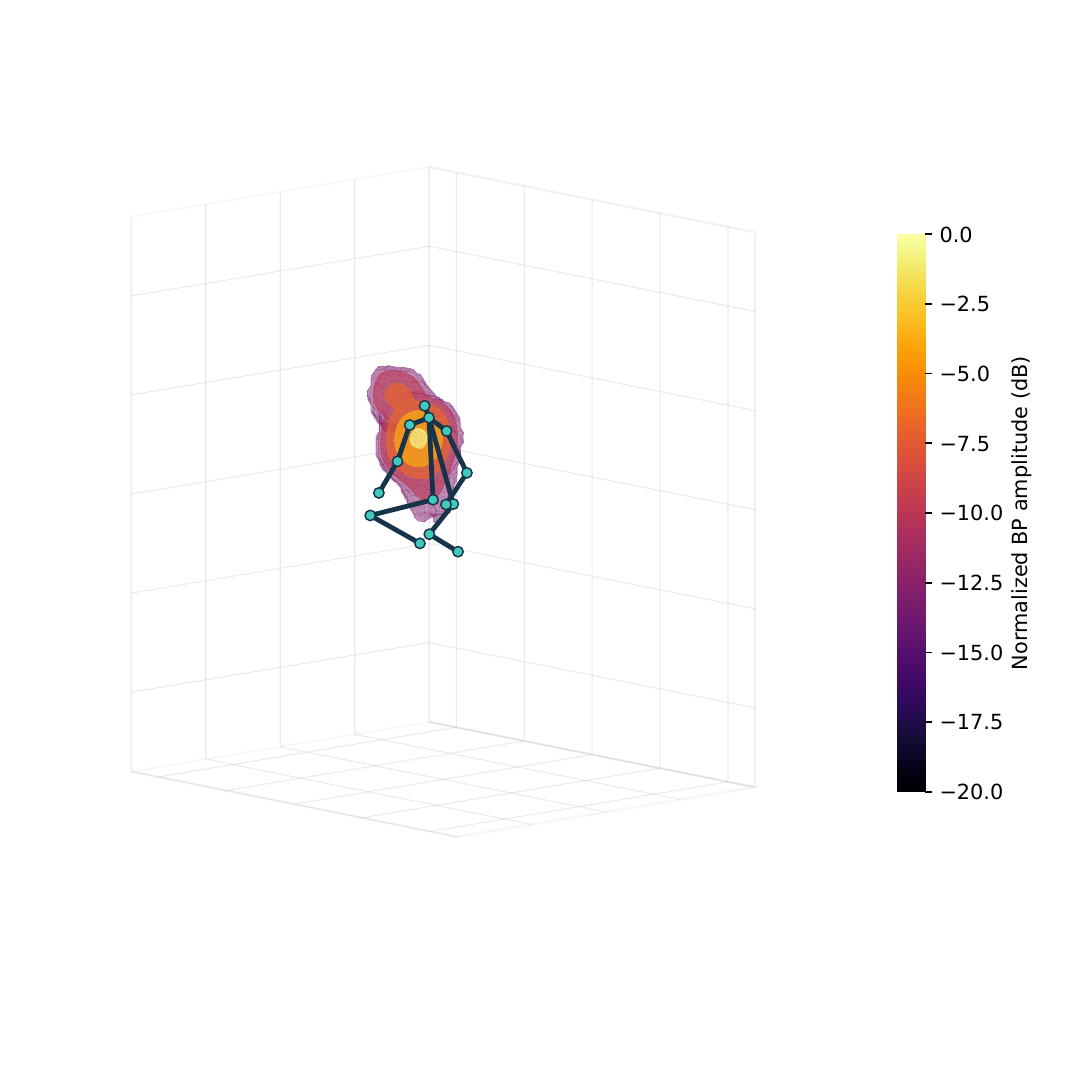}\\[4pt]
BP + 3D CNN$^{\dagger}$ &
\qualitativepanel{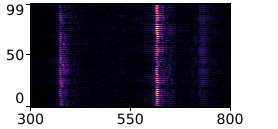} &
\qualitativepanel{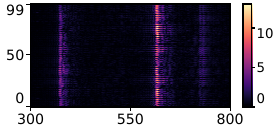} &
\qualitativeposepanel{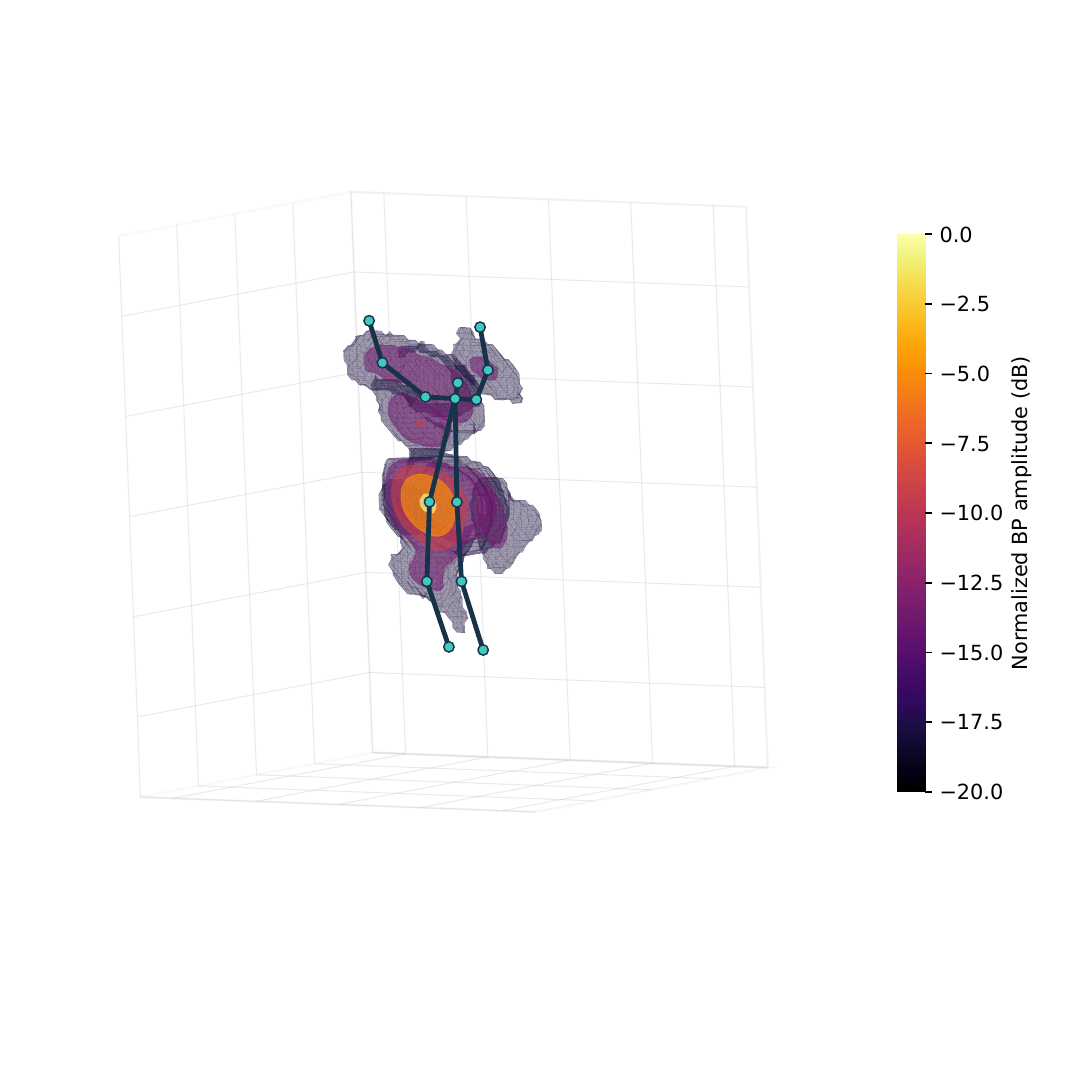}\\[4pt]
RFMamba$^{\dagger}$ &
\qualitativepanel{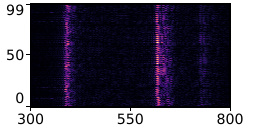} &
\qualitativepanel{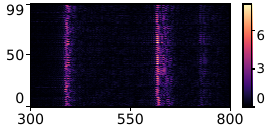} &
\qualitativeposepanel{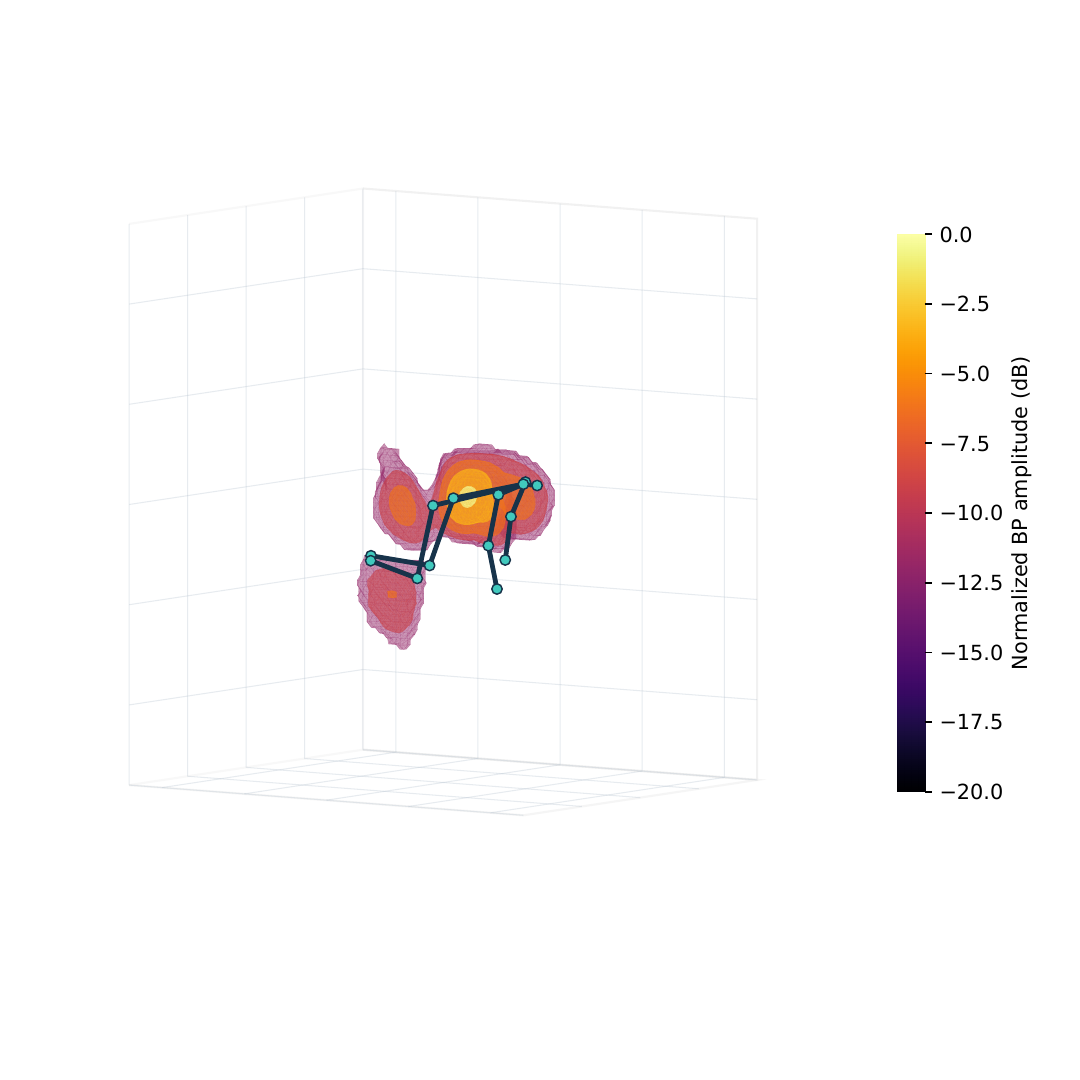}
\end{tabular}
\endgroup
\caption{\bipara{fig_qualitative}{Echo generation and downstream pose estimation examples. (a,b) Measured and generated echo amplitudes, with fast-time indices horizontally and MIMO channels vertically. (c) Skeletons predicted by the corresponding estimators, overlaid on 3D BP reconstructions of the generated echoes.}}
\label{fig:qualitative}
\end{figure*}

\bipara{ex13}{Figure~\ref{fig:qualitative} presents qualitative results. In these examples, the generated echoes closely reproduce the dominant amplitude bands and their variation across MIMO channels. The predicted skeletons also align well with the body structure visible in the 3D BP reconstructions of the generated echoes.}

\subsection{Transfer from visual pose datasets}\label{sec:transfer}
\bipara{ex08}{We use 3D keypoint annotations from CMU Panoptic, AIST++, and FreeMan, mapped to our 14-joint skeleton~\cite{Joo_2015_ICCV,Li_2021_ICCV,Wang_2024_CVPR}. With the generator frozen, poses from these public datasets provide generated echo and pose pairs for estimator training.}

\bipara{ex09}{Table~\ref{tab:transfer} evaluates the same trained estimator in two domains. Source-generated MPJPE uses echoes generated from test poses in the optical datasets and their labels to assess pose consistency within the generated pairs. Real-target MPJPE uses measured radar test echoes and their paired labels to assess whether the transferred supervision remains useful for real radar inputs. We compare BP + 3D CNN with the single-frame spatial adaptation RF-P3D-SF. We exclude RadarFormer and RFMamba because they require temporal inputs, whose validity is not guaranteed in our transfer protocol for public pose annotations.}

\bipara{ex10}{Table~\ref{tab:transfer} shows that BP + 3D CNN achieves lower error than RF-P3D-SF for every pose source in both test domains. AIST++ yields the lowest measured-test errors for both estimators. One explanation is that AIST++ mainly contains dance poses with a wide range of motion, providing useful coverage of body configurations for learning features that transfer to the radar test set. For BP + 3D CNN, training with transferred data gives performance close to the real-only baseline or better, depending on the pose source (Table~\ref{tab:augmentation}). Although the gains vary across sources, these results demonstrate that public visual pose annotations can provide effective supervision for radar pose estimation, making their scale and pose coverage available to radar learning.}

\subsection{Effect of alignment pretraining}\label{sec:ablations}

\bipara{ex11}{To isolate the effect of alignment pretraining, we initialize SkeleCoder randomly and train it jointly with diffusion, leaving all other settings unchanged. Echoes generated by each variant augment the measured training set for all four estimators under the same augmentation settings. Table~\ref{tab:ablation} reports MPJPE on the measured test set.}

\begin{table}[t]
\centering
\caption{\bipara{tab03}{Effect of alignment pretraining on radar data augmentation. MPJPE (mm) is evaluated on the measured test set. $\Delta$ is the increase after removing pretraining.}}
\label{tab:ablation}
\small
\setlength{\tabcolsep}{4pt}
\begin{tabular}{@{}lccc@{}}
\toprule
Estimator & \shortstack{With alignment} & \shortstack{Without alignment} & $\Delta$\\
\midrule
RadarFormer & $38.5$ & $92.7$ & $+54.2$\\
RF-Pose3D$^{\dagger}$ & $39.1$ & $112.5$ & $+73.4$\\
BP + 3D CNN$^{\dagger}$ & $36.6$ & $107.4$ & $+70.8$\\
RFMamba$^{\dagger}$ & $51.3$ & $96.4$ & $+45.1$\\
\bottomrule
\end{tabular}
\end{table}

\bipara{ex12}{Removing pretraining increases MPJPE by 45.1--73.4 mm across all four estimators (Table~\ref{tab:ablation}). The imaging methods RF-Pose3D and BP + 3D CNN show larger increases, at 73.4 and 70.8 mm, than the direct echo models RadarFormer and RFMamba, at 54.2 and 45.1 mm. All four variants also perform worse than training on measured data alone (Table~\ref{tab:augmentation}). Joint training from random initialization therefore does not recover the augmentation benefit of alignment pretraining within the same diffusion budget. The larger degradation of imaging methods may be related to their reconstruction step, where echo errors can disturb coherent focusing and intensity conversion discards phase information available to direct echo estimators. These results support prealigning skeleton and radar features before joint diffusion training.}

\section{Conclusion}\label{sec:conclusion}
\bipara{co01}{We presented \methodname{}, which turns 3D pose annotations into paired complex MIMO radar echoes for learning. MIMO-VAE and the pretrained SkeleCoder provide radar and pose representations for conditional diffusion. The same generator improves four radar pose estimators through data augmentation and transfers annotations from three public visual pose datasets into useful radar supervision. These results connect radar learning to existing visual data resources and suggest a way to expand training data with less dependence on synchronized acquisition. The current generator operates on individual poses and does not explicitly model temporal consistency. Extending it to coherent motion sequences could support larger radar training datasets for human sensing across a wider range of activities.}

% Official Author Guidelines: main text excludes the LLM statement and references.
% Flush all main-text floats before measuring the eight-page budget.
\clearpage
\section*{LLM Usage Statement}
\label{sec:llm_statement}
\bipara{llm01}{OpenAI Codex was the only LLM tool used in this work. It assisted with LaTeX typesetting, grammar checking and sentence polishing, code development, and color scheme design for figures.}

\end{document}